# Hyperparameters Optimization in Deep Convolutional Neural Network / Bayesian Approach with Gaussian Process Priors

Pushparaja Murugan

*School of Mechanical and Aerospace Engineering,*

*Nanyang Technological University, Singapore 639815.*

(pushpara001@e.ntu.edu.sg )

## Abstract

Convolutional Neural Network is known as ConvNet have been extensively used in many complex machine learning tasks. However, hyperparameters optimization is one of a crucial step in developing ConvNet architectures, since the accuracy and performance are totally reliant on the hyperparameters. This multilayered architecture parameterized by a set of hyperparameters such as the number of convolutional layers, number of fully connected dense layers & neurons, the probability of dropout implementation, learning rate. Hence the searching the hyperparameter over the hyperparameter space are highly difficult to build such complex hierarchical architecture. Many methods have been proposed over the decade to explore the hyperparameter space and find the optimum set of hyperparameter values. Reportedly, Gird search and Random search are said to be inefficient and extremely expensive, due to a large number of hyperparameters of the architecture. Hence, Sequential model-based Bayesian Optimization is a promising alternative technique to address the extreme of the unknown cost function. The recent study on Bayesian Optimization by Snoek in nine convolutional network parameters is achieved the lowerest error report in the CIFAR-10 benchmark. This article is intended to provide the overview of the mathematical concept behind the Bayesian Optimization over a Gaussian prior.

## 1  Introduction

Model accuracy has been dramatically influenced by the selection of a set of hyperparameters, in some cases changing it from 1 to 95 [1] [2]. Over the past few years, a variety of algorithms are developed for optimizing the hyperparameters values such as grid search, random search [3], model-based approaches using random forests [4] and sequential model-based optimization [5]. The widely used optimization algorithm is grid search method. In this grid search method, the model is trained over a range of the set of hyperparameter, and the values give the best performance on the network on the cross-validation data is selected. Also, this method involves training the network with manually selected sub-set of hyperparameters and explore the defined hyperparameter space which is considered to be ineffective and extremely time-consuming. This method is severely influenced by the curse of dimensionality as the number of hyperparameters increase. The random search method is the effective alternative of grid search to explore the randomly sampled hyperparameters in the hyperparameter space. Its also time-consuming when involves a large number of hyperparameters in the search space. One of the most powerful strategies to optimize the hyperparameter is Sequential model-based optimization. In this technique involves constructing a probabilistic model to the data to determine the most promising point to evaluate. Hyperparameter optimization requires optimizing an unknown black box function. In such computationally expensive situation , Bayesian optimization technique is an efficient and powerful heuristic to optimize the function [6] [7]. Sequential model-based Bayesian Optimization are studied by many researchers and are well established global optimization strategy for unknown noisy function [8]. Bayesian optimization constructs a probabilistic surrogate model to define the distribution over the unknown black box function, and a proxy optimization is performed to seek the next location to evaluate where the posterior distribution is developed based on conditioning on the previous evaluations. Acquisition function is applied to the posterior mean and variance to express the trade off between exploration and exploitation. Most commonly, Gaussian Process is used in the construction of the distribution over objective function because of their flexibility, well-calibrated uncertainty, and analytic properties [9]. Bayesian Optimization is shown the better performance than grid search and random search and outperformed the state of art performance on several machine learning task [10] [11]. Swersky et al. investigation on Mutiltak Bayesian Optimization showed that the knowledge transformation between correlated tasks where they tried to determine the best configuration for the dataset by evaluating manually chosen sub-sets [12]. Nickson et at, estimated configuration performance of larger dataset by the evaluation based on several training on small random manually chosen subsets [13]. Bergstra et al., showed that complex vision architecture such as face matching verification & identification and object recognition is achieved the state of the art performance with 238 hyperparameters [14]. Also, many studies suggested that automated hyperparameter optimization is significantly improved the model performance for wide range of problems [15] [16] [17].

The purpose of this article is to provide the fundamental theoretical concepts and mathematical formulation of Bayesian approach over the Gaussian Process prior to determining the extreme of objective cost functions. This article is developed as a continuation of our previous study [18] [19].

## 2 Bayesian Optimization

Optimzation is a mathematical technique to find the maxmium or minium value of a objective function of several variable subjected to set of constraints. For given a real valued function $f : \chi \subset \mathbb{R}^n \to \mathbb{R} = \{f(x) : x \in \chi\}$ in a defined a set of $\chi$, has the maxmium on the set $\chi$ at a point $x_\ell \in \chi$ provided that $f(x_\ell \geq f(x))$ for all $x \in \chi$, than the maximum value $f(x_\ell)$ is expressed as,

$$max\{f(x) : x \in \chi\} \qquad \text{(Eq. 2.1)}$$

where $f(x_\ell)$ is the maximum of the function and $x_\ell$ is the maximizer of the function. If a point $x_\ell$ is the maximum value of a function $f(x)$, the same point $x_\ell \in \chi$ is the minimum value of $-f(x)$ function provided that $f(x_\ell \leq f(x))$. Here, the optimization is considered to be minimization and $x_\ell$ is the minimizer of the function. However, the black box function does not have any explicit expression for objective functions. Hence, the evaluation must be carried out by querying point $x \in \chi$ and gaining the corresponding response. Bayesian optimization is one of the powerful strategy used in determining the extreme of an objective function $f(x)$ on a bounded set of $\chi \in \mathbb{R}^n$. Bayesian optimization is extremely useful in determining the extreme in black box function that does not have any expression or derivatives or in non-convex function. Bayesian optimization constructs a probabilistic model for the objective function and exploits the decision about the next promising location of the function while integrating out the uncertainty. In finding the next location to evaluate, selecting the prior over the function and acquisition functions are important that can be obtained by incorporating the prior belief about the objective function and the trade-off of exploration & exploitation.

**Definition 1.** *Bayes theorem states that the posterior probability of a model $\mathcal{M}$ given observation $\mathcal{D}$ is proportional to the likelihood of $\mathcal{D}$ given $\mathcal{M}$ multiplied by the prior probability of $\mathcal{M}$.*

$$P(\mathcal{M}|\mathcal{D}) \propto P(\mathcal{D}|\mathcal{M})P(\mathcal{M}) \qquad \text{(Eq. 2.2)}$$

The prior is represent the belief about the space of possible objective function in Bayesian Optimization. If $f(x_\ell)$ is the observation of the objection function at $i^{th}$ sample $x_\ell$, for an accumulated observation $\mathcal{D}_{1:t} = \{x_{1:t}, f(x_{1:t})\}$, the prior distribution is combined with the likelihood function $P(\mathcal{D}_{1:t}|f(x_{1:t}))$, the posterior distribution can be written as,

$$P(f(x_{1:t})|\mathcal{D}_{1:t}) \propto P(\mathcal{D}_{1:t}|f(x_{1:t}))P(f(x_{1:t})) \qquad \text{(Eq. 2.3)}$$

This posterior distribution captures the updated beliefs about the objective function. Bayesian optimization utilizes the acquisition function to find the next location point $x_{t+1} \in \mathbb{R}^d$ for sampling. This automatic representation of trade-off of explorations and exploitations help to minimize the number of location points on the objective function to evaluate. This technique can be used in the objective function with multiple local extremes. Evaluation of next point is carried out by computing the maximum in the acquisition function which can be done by accounting the mean and covariance of the predictions. Then the objective is sampled at $\arg\max x$ of the functions. The GP is updated, and the process is repeated. The prior and observation are used to define the posterior distribution over the spaces of the objective functions where the informative priors are describing the attributes of the functions include smoothness and extreme even when the function itself is not known.

## 2.1  Gaussian process over a function

Gaussian process is a stochastic approach to modelling the observed data over time or space or time and space that is determined by mean and covariance. Gaussian Processes used to introduce the correlation between the data vector by constructing joint probability of that vector.

**Definition 2.** *A Gaussian Process is a collection of random variables, any finite number of which have (consistent) joint Gaussian distributions.*

A Gaussian process is the generalization of Gaussian probability distribution where the probability distribution describes random variables(scalar or vector) and the properties are governed by the stochastic process. For any given choice of distinct long input vector $x_\ell = \{x_1, x_2, \ldots x_n\}, i \in \chi$ for all $i$, the output is $f(x_\ell) = \{f(x_1), f(x_2) \ldots f(x_n)\}$ and is gaussian distributed. The Gaussian distribution over a vector $f(x_\ell)$ has multivariate normal distribution with mean vector $\mu = \mathbb{E} f(x_\ell)$ and covariance matrix $\gamma = Cov(f(x_\ell), f(x_\ell))$,

$$f(x_\ell) \sim \mathcal{GP}(\mu, \gamma) \qquad \text{(Eq. 2.4)}$$

If the function is distributed as Gaussian Processes with mean function $\mu : \chi \to \mathbb{R}$ and covarianace or kernel function $\gamma : \chi \times \chi \to \mathbb{R}$, for any given input $x_\ell = \{x_1, x_2, \ldots x_n\} \in \chi$, the output function $f(x_\ell) = \{f(x_1), f(x_2) \ldots f(x_n)\}$ is gaussian distrubited with mean $\mu_i = [\mu(x_1), \mu(x_2), \mu(x_3) \ldots \mu(x_n)]$ and $n \times n$ covariance or kernel matrix $\gamma(x_\ell, x'_\ell)$, We say,

$$f(x_\ell) \sim \mathcal{GP}(\mu(x), \gamma(x, x')) \qquad \text{(Eq. 2.5)}$$

where, the mean function is,

$$\mu(x_\ell) = \mathbb{E}[f(x_\ell)] \qquad \text{(Eq. 2.6)}$$

The covariance function is,

$$\gamma(x_\ell, x'_\ell) = \mathbb{E}[(f(x_\ell) - \mu(x_\ell))(f(x'_\ell) - \mu(x'_\ell))^T] \qquad \text{(Eq. 2.7)}$$

## 2.2 Computing Posterior

The Gaussian Process distribution over a function is used as prior to Bayesian inference where the prior describes the properties of the function such as smooth, quadratic and extreme. Computing posterior can be used to predict unseen test data. Computation of posterior of the function can be done by finding the joint distribution of $(f(x_\ell), f(x_\ell)_*)$ and using the conditional rule for gaussian to compute the conditional distribution of $(f(x_\ell), f(x_\ell)_*)$. If $f(x_\ell)$ is the known set of function values of training data and $f(x_\ell)^*$ is the function values to the set of inputs $x_\ell^*$, then the joint distribution of the function is written as, assume the the prior mean function $\mu(x_\ell) = 0$, the posterior distribution process is given as [20],

$$f(x_\ell) \sim \mathcal{GP}(m_{\mu(x_\ell)}, k_{\gamma(x_\ell)}), \qquad \text{(Eq. 2.8)}$$

$$\mu(x_\ell) = \mu(x_\ell) + \gamma(x_\ell, x_\ell)^T \gamma^{-1}(f(x_\ell) - m(x_\ell)) \qquad \text{(Eq. 2.9)}$$

$$\gamma(x_\ell, x'_\ell) = \gamma(x_\ell, x'_\ell) - \gamma(x_\ell . x_\ell)^T - \gamma^{-1} \gamma(x_\ell, x'_\ell) \qquad \text{(Eq. 2.10)}$$

where $\gamma(x_\ell, x_\ell)$ is represent the covariances matrix of training data.

$$\begin{bmatrix} f(x_\ell^*) \\ f(x_i) \end{bmatrix} \sim \mathcal{N}\left( \begin{bmatrix} 0 \\ 0 \end{bmatrix}, \begin{bmatrix} \gamma(x_\ell^*, x_\ell^*) & \gamma(x_\ell^*, x_\ell)^T \\ \gamma(x_\ell^*, x_\ell) & \gamma^{**} \end{bmatrix} \right) \qquad \text{(Eq. 2.11)}$$

where,

$$\gamma(x_\ell^*, x_\ell) = \begin{bmatrix} \gamma(x_\ell^*, x_{i1}) \\ \gamma(x_\ell^*, x_{in}) \end{bmatrix} \quad \text{(Eq. 2.12)}$$

Using the conditional rule to obtained the posterior for $f(x_\ell^*)$ is guassian,

$$f(x_\ell^*)|f(x_\ell) \sim \mathcal{N}(\gamma(x_\ell^*, x_\ell)^T \gamma_{x_\ell x_\ell}^{-1} f(x_\ell), k(x_\ell^*, x_\ell *) + k(x_\ell^*, x_\ell)^T \gamma_{x_\ell x_\ell}^{-1} \gamma(x_\ell^*, x_\ell)) \quad \text{(Eq. 2.13)}$$

where, $\mu(x_\ell)$ and $\gamma$ are represent the mean and covariance of the distribution of training case and $\mu(x_\ell)^*$ and $\gamma^{**}$ is represent the mean of the distribution of the test cases. The Posterior mean $\mathbb{E}(f(x_\ell^*)|f(x_\ell))$ is represent as a linear combination of kernel or covariance function values,

$$\mathbb{E}(f(x_\ell^*|f(x_\ell)) = \sum_{i=1}^{n} \alpha_i \gamma(x_\ell^*, x_\ell) \quad \text{(Eq. 2.14)}$$

for $\alpha_i = \gamma_{x_\ell x_\ell}^{-1} f(x_\ell)$. The mean of the distribution can be computed by solving $\gamma \alpha_i = f(x_\ell)$. Similarly, The linear combination of the obsevered values of the function,

$$\mathbb{E}(f(x_\ell^*)|f(x_\ell)) = \sum_{i=1}^{n} \beta_i f(x_\ell) \quad \text{(Eq. 2.15)}$$

where, $\beta_i = \gamma(x_\ell^*, x_\ell)^T \gamma^{-1}$ For non-zero mean prior,

$$\begin{bmatrix} f(x_\ell^*) \\ f(x_i) \end{bmatrix} \sim \mathcal{N}\left( \begin{bmatrix} \mu(x_\ell^*) \\ \mu(i) \end{bmatrix}, \begin{bmatrix} \gamma(x_\ell^*, x_\ell^*) & \gamma(x_\ell^*, x_\ell)^T \\ \gamma(x_\ell^*, x_\ell) & \gamma^{**} \end{bmatrix} \right) \quad \text{(Eq. 2.16)}$$

In case of noisy observation of the objective function at $x_\ell$ is $y_i = f(x_\ell) + \epsilon_i$ where $\epsilon_i \sim \mathcal{N}(0, \sigma_{noise}^2)$, assume that the $\epsilon$ is zero mean guassian noise, then the posterior can be computed by finding the joint distribution of $[f(x^*), y(x_1) \ldots y(x_n)]$ and use conditional distribution as before. If the prior mean function is $m(x) = 0$,

$$\begin{bmatrix} f(x_\ell^*) \\ y(x_\ell) \end{bmatrix} \sim \mathcal{N}\left( \begin{bmatrix} \mu(x_\ell^*) \\ \mu(i) \end{bmatrix}, \begin{bmatrix} \gamma(x_\ell^*, x_\ell^*) & \gamma(x_\ell^*, x_\ell)^T \\ \gamma(x_\ell^*, x_\ell) & \gamma^{**} + \sigma^2 \mathcal{I} \end{bmatrix} \right) \quad \text{(Eq. 2.17)}$$

The covariance matrix appears with $\sigma^2$ term on diagonal in noisy observation case. Noise is independent of different observation and function. Hence, there is no covariance between the noise and the function. The posterior on $f(x_\ell^*)$ is,

$$f(x_\ell^*)|y(x_\ell) \sim \mathcal{N}(\gamma(x_\ell^*, x_\ell)^T (\gamma_{x_\ell x_\ell} + \sigma^2 \mathcal{I})^{-1} y(x_\ell), k(x_\ell^*, x_\ell^*) + \sigma^2 + \gamma(x_\ell^*, x_\ell)^T (\gamma_{x_\ell x_\ell} + \sigma^2 \mathcal{I})^{-1} \gamma(x_\ell^*, x_\ell)) \quad \text{(Eq. 2.18)}$$

## 2.3 Covariance functions

The covariance function $\gamma(x_\ell, x_\ell)$ for the GPs is the crucial ingredients since it defines the properties nearness or similarity such as of the evaluation points. A typical kernel or covariance functions are generalized by hyperparameters. However, Single hyperparameter $\theta$ is added to generalize the isotropic covariance.

$$\gamma(x_\ell, x_\ell') = \sigma_f^2 \exp\{-\frac{1}{2l^2}(x_\ell - x_\ell')^2\} \quad \text{(Eq. 2.19)}$$

For anisotropic models, the squared exponential kernel with a vector of automatic relevance determination (ARD) hyperparameters $\theta$

$$\gamma(x_\ell, x_i') = \exp(-\frac{1}{2}(x_\ell - x_i')^T diag(\theta)^{-2}(x_\ell - x_\ell')) \tag{Eq. 2.20}$$

where $diag(\theta)$ is represent the diagonal matrix. Another important kernel for BO is the Matern kernel that incorporates the smoothness $\varsigma$ for the flexibility on modelling the functions [21].

$$\gamma(x_\ell, x_i') = \frac{1}{2^{\varrho-1}\gamma(\varrho)}(2\sqrt{\varrho}||x_\ell - x_i'||)^\varrho H_\varrho(2\sqrt{\varrho}||x_\ell - x_i'||) \tag{Eq. 2.21}$$

where $\gamma$ and $H_\varrho$ are represent the Gamma and Bessel function respectively.

Matern kernel reduces to squared exponential kernel, if $\varrho \to \infty$ and reduces to unsquared expoential kernel $\varrho \to 0.5$

## 2.4 Acquisition Functions

The objective of the acquisition function is to provide the guidance to locate the optima of function. High values of acquisition are indicated the high values of the function since the prediction is high at that point or uncertainty is high, or both are present in that point. The next point to query in the function is determined by maximizing the acquisition function. Maximized the acquisition function is then used to perform the evaluation $\arg\max_x u(x|\mathcal{D})$ at promising point $x_\ell$ on the function where $u$ is represent the acquisition function.

## 2.5 Probability of Improvement and Expected Improvement

Expected improvement based acquistion function is to compute the probability of improvement with respect to current maxmium [22] $f(x^+)$, where, where $x_\ell^+ = \arg\max_{x_\ell \in x_{1:t}} f(x_\ell)$, so that,

$$PI(x_\ell) = P(f(x_\ell) \geq f(x_\ell^+)) \tag{Eq. 2.22}$$

Maximum probability of improvement is,

$$= \phi\left(\frac{\mu(x_\ell) - f(x_\ell^+)}{\sigma(x_\ell)}\right) \tag{Eq. 2.23}$$

where, $\phi$ is present the Cumulative Distribution Function. Maximum probability of improvement is purely based on exploitation. To consider the promising points which have a higher probability of greater than $f(x^+)$, a trade-off parameter is added $\xi \geq 0$. Hence the maximum probability of improvement is,

$$PI(x_\ell) = P(f(x_\ell) \geq f(x_\ell^+) + \xi) \tag{Eq. 2.24}$$

$$= \phi\left(\frac{\mu(x_\ell) - f(x_\ell^+) - \xi}{\sigma(x_\ell))}\right) \tag{Eq. 2.25}$$

The value of $\xi$ is usually high at the beginning of the optimization to obtain the exploration and decreased towards zero. Minimizing the expected improvement from the current maximum $f(x_\ell^*)$ at a new location point, probability of improvement and magnitude of the improvement is considered, Hence the more satisfying alternative acquisition function is,

$$x_{t+1} = \arg\max_x \mathbb{E}(||f_{t+1}(x_\ell) - f(x_\ell^*)|| \ |\mathcal{D}_{1:t}) \quad \text{(Eq. 2.26)}$$

$$\arg\max_x \int ||f_{t+1}(x_\ell) - f(x_\ell^*)|| P(f_{t+1}|\mathcal{D}_{\infty:\sqcup}) df_{t+1} \quad \text{(Eq. 2.27)}$$

However, this decision process is considered only the choices of one step ahead. In order to consider the two-step ahead,

$$x_{t+1} = \arg\max_{x_\ell} \mathbb{E}\left(\min_{x_\ell'} \mathbb{E}(||f_{t+2}(x_\ell') - f(x_\ell^*)|| \ \mathcal{D}_{t+1})|\mathcal{D}_{t+1}\right) \quad \text{(Eq. 2.28)}$$

Because of the computational expense of the method, alternative of maximizing the expected improvement is proposed [23] and is given by,

$$\mathcal{I}(x_\ell) = max\{0, f_{t+1}(x_\ell) - f(x_\ell^+)\} \quad \text{(Eq. 2.29)}$$

where, $\mathcal{I}(x)$ is positive if the prediction is high, else $\mathcal{I}(x) = 0$. The next location of point is determined by maximizing the expected improvement and is expressed as,

$$x_\ell = \arg\max_{x_\ell} \mathbb{E}(max\{0, f_{t+1}(x_\ell) - f(x_\ell^+)\}|\mathcal{D}_t) \quad \text{(Eq. 2.30)}$$

The likelihood of improvement $\mathcal{I}$ is computed from normal density function,

$$\frac{1}{\sqrt{2\pi}\sigma(x_\ell)} e^{\left(-\frac{(\mu(x_\ell) - f(x_\ell^+) - I)^2}{2\sigma^2(x_\ell)}\right)} \quad \text{(Eq. 2.31)}$$

The expected improvment can be computed by integrating over likelihood improvement function [7],

$$\mathbb{E}(\mathcal{I}) = \int_{\mathcal{I}=0}^{\mathcal{I}=\infty} \mathcal{I} \frac{1}{\sqrt{2\pi}\sigma(x_\ell)} e^{\left(-\frac{(\mu(x_\ell) - f(x_\ell^+) - I)^2}{2\sigma^2(x_\ell)}\right)} d\mathcal{I} \quad \text{(Eq. 2.32)}$$

$$\mathbb{E}(\mathcal{I}) = \sigma(x_\ell)\left[\frac{\mu(x_\ell) - f(x_\ell^+)}{\sigma(x_\ell)} \phi\left(\frac{\mu(x_\ell) - f(x_\ell^+)}{\sigma(x_\ell)}\right) + \psi\left(\frac{\mu(x_\ell) - f(x_\ell^+)}{\sigma(x_\ell)}\right)\right] \quad \text{(Eq. 2.33)}$$

$$\mathbb{E}\mathcal{I}(x_\ell) = \begin{cases} (\mu(x_\ell) - f(x_\ell^+))\phi(Z_i)\sigma(x_\ell)\psi(Z_i)) & if \ \sigma(x_\ell) > 0 \\ 0 & if \ \sigma(x_\ell) = 0 \end{cases} \quad \text{(Eq. 2.34)}$$

$$Z_i = \frac{\mu(x_\ell) - f(x_\ell^+)}{\sigma(x_\ell)} \quad \text{(Eq. 2.35)}$$

where $\phi$ and $\psi$ denote the Probability Density Function and Cumulative Distribution Functions of normal distribution.

## 2.6 Explorations & exploitations trade off

The trade-off between exploiting and exploring is balanced by the expected improvement with respect to the distribution of GPs. On exploring, the points have the largest surrogate variance should be considered where on exploiting, the points have the highest surrogate mean is to be considered. Hence, the expected improvement $\mathbb{E}I$ regulate the global and local optimization trade-off,

$$\mathbb{EI}(x_\ell) = \begin{cases} (\mu(x_\ell) - f(x_\ell^+) - \epsilon)\phi(Z_i)_\sigma(x_\ell)\psi(Z_i)) & if \ \sigma(x_\ell) > 0 \\ 0 & if \ \sigma(x_\ell) = 0 \end{cases} \quad \text{(Eq. 2.36)}$$

where,

$$Z_i = \begin{cases} \frac{\mu(x_\ell) - f(x_\ell^+) - \epsilon}{\sigma(x_\ell)}) & if \ \sigma(x_\ell) > 0 \\ 0 & if \ \sigma(x_\ell) = 0 \end{cases} \quad \text{(Eq. 2.37)}$$

where $\epsilon \geq 0$

## 2.7 Confidence bound criteria

Cox has developed a Sequential Design for Optimization that selects the point to evaluate based on the lower bound confidence of the prediction [24]. The lower bound confidence is given as,

$$\mathcal{LCB}(x_\ell) = \mu(x_\ell) - \Upsilon\sigma(x_\ell) \quad \text{(Eq. 2.38)}$$

where $\Upsilon \geq 0$. For maxmization, the upper confidence bound is given as,

$$\mathcal{UBC}(x_\ell) = \mu(x_\ell) + \Upsilon\sigma(x_i) \quad \text{(Eq. 2.39)}$$

# 3 Conclusion

Hyperparameter optimization is one of the important steps in developing ConvNet architecture without compromising the computational expensive. Though the Grid search and Random search are widely used in many machine learning tasks, they are ineffective and time-consuming as the number of hyperparameters is increased. Hence, a well-sophisticated optimization algorithm is needed to be addressed. In that connection, Bayesian Optimization over GPs is promising candidate for optimizing a large number of hyperparameters. In this article, theories and mathematical concepts of Bayesian Optimization approach are explained elaborately.

# Aknowledgement

We thank Dr.Eric Brochu for making his PhD thesis available in internet.